\documentclass[10pt]{article} 

\usepackage{arxiv}

\usepackage{amssymb}            
\usepackage{mathtools}          
\usepackage{mathrsfs}           
\usepackage{graphicx}           
\usepackage{subcaption}         
\usepackage[space]{grffile}     
\usepackage{url}                
\usepackage{lipsum}             
\usepackage{microtype}
\usepackage{booktabs}
\usepackage{subcaption}
\usepackage{wrapfig}
\usepackage{natbib}
\usepackage{hyperref}

\setcitestyle{authoryear,open={(},close={)}}

\usepackage{authblk}

\newcommand{\methodname}{Multi-Agent Craftax}
\newcommand{\methodabbrev}{MAC}

\title{An Efficient Open World Environment for Multi-Agent Social Learning}
\author[1]{Eric Ye}
\author[1]{Ren Tao}
\author[1]{Natasha Jaques}

\affil[]{Paul G. Allen School of Computer Science and Engineering, University of Washington}
\affil[]{\textit {\{ericy4,taocla17,nj\}@cs.washington.edu}}

\begin{document}
\maketitle

\begin{abstract}
    Many challenges remain before AI agents can be deployed in real-world environments. However, one virtue of such environments is that they are inherently multi-agent and contain human experts. Using advanced social intelligence in such an environment can help an AI agent learn adaptive skills and behaviors that a known expert exhibits. While social intelligence could accelerate training, it is currently difficult to study due to the lack of open-ended multi-agent environments. In this work, we present an environment in which multiple self-interested agents can pursue complex and independent goals, reflective of real world challenges. 
    This environment will enable research into the development of socially intelligent AI agents in open-ended multi-agent settings, where agents may be implicitly incentivized to cooperate to defeat common enemies, build and share tools, and achieve long horizon goals. In this work, we investigate the impact on agent performance due to social learning in the presence of experts and implicit cooperation such as emergent collaborative tool use, and whether agents can benefit from either cooperation or competition in this environment.

\end{abstract}
\section{Introduction}

The real world is inherently multi-agent. For AI systems ranging from autonomous cars to household and digital assistants to be useful to people, they require social intelligence to effectively navigate human multi-agent environments. 
Despite additional challenges like coordinating with other agents, multi-agent systems also present new opportunities.
As intelligent (human) agents complete their independent tasks in the real world, they demonstrate highly effective behaviors honed by a lifetime of experience \citep{rendell2010copy}. If AI agents could effectively leverage this information, they could not only learn to rapidly acquire complex skills, but also adapt online to environmental changes, addressing fundamental challenges in AI relating to learning and generalization.



In this paper, we introduce \textbf{\methodname{} (\methodabbrev)}, an open-ended MARL environment designed to test the social intelligence of AI agents in a highly performant manner. Our environment is capable of investigating questions related to social learning, collaborative tool use, the emergence of complex skills, and how these factors are affected by incentives to cooperate or compete. Built on the high-performance, JAX-based Craftax environment \citep{craftax2024}, \methodabbrev{} is capable of running 100 million training steps in less than one hour on a single GPU. 
In this dynamic environment, multiple agents can pursue independent goals, which each require complex, long-horizon planning in a large, partially-observable, sparse reward environment. Importantly, the goals share underlying subtasks, such as tool crafting, so agents with social learning and cooperation skills can more effectively achieve their goals. Thus, 
\methodabbrev{} is designed to facilitate research into methods that improve agents' abilities  to learn, interact, and adapt in the presence of other intelligent agents. 

\begin{figure}[t]
    \centering
    \begin{subfigure}{0.3\linewidth}
        \includegraphics[width=\linewidth]{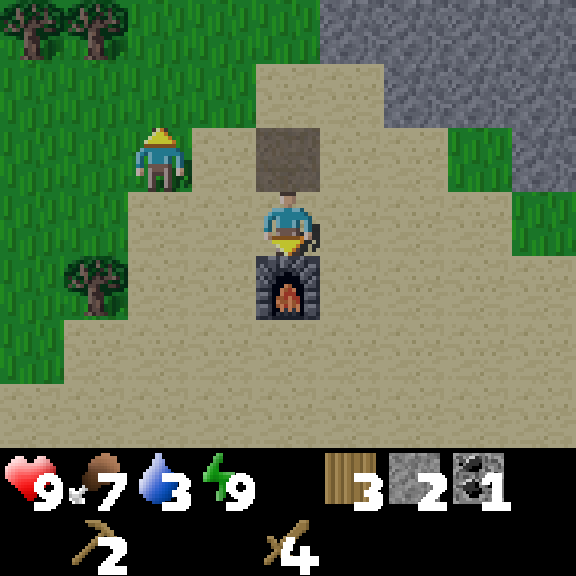}
    \end{subfigure}%
    ~
    \begin{subfigure}{0.3\linewidth}
        \includegraphics[width=\linewidth]{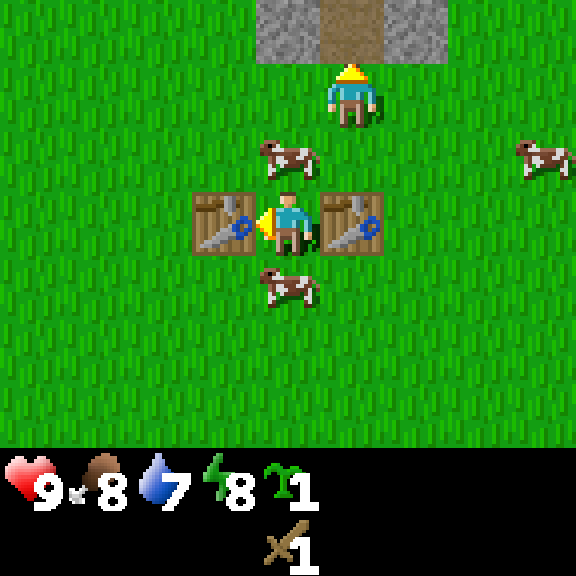}
    \end{subfigure}%
    ~
    \begin{subfigure}{0.3\linewidth}
        \includegraphics[width=\linewidth]{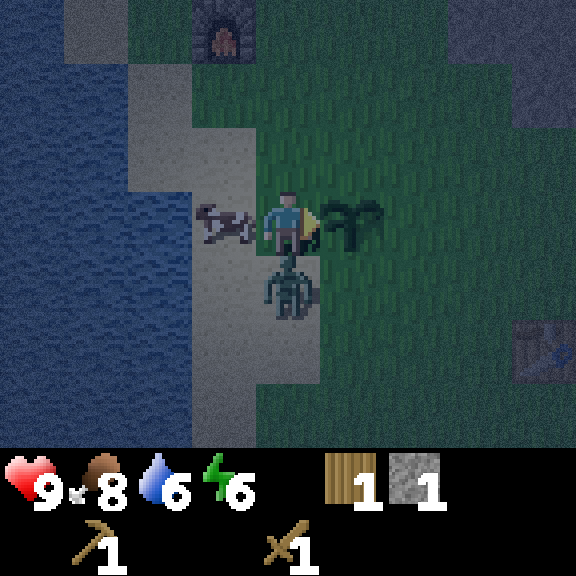}
    \end{subfigure}
    \caption{Sample pixel observations from a single agent in \methodabbrev{}. We compare the performance of agents trained with others to their single-agent counterparts to study whether the presence of other agents accelerates learning. We find that agents do not benefit from existing social learning methods, pointing to the need for more sophisticated algorithms. Yet, agents use tools crafted by other agents 90\% of the time, highlighting implicit cooperation in the shared environment.}%
    \label{fig:game}
\end{figure}

\textbf{Social Learning}---learning how to learn from other agents---could improve generalization of AI agents by enabling them to rapidly acquire new skills from experts present in their environment \citep{ndousse2021emergent,filos2021psiphi,bhoopchand2023learning}. Consider an office robot tasked with processing trash. Initially unaware of the trash categorizations and bin locations, it can rapidly learn these by observing where workers throw trash into which bins.
Exploration is costly, creating conditions where social learning is especially beneficial \citep{laland2004social}. In this benchmark, we use recently proposed metrics~\citep{bhoopchand2023learning}, to measure the performance of state-of-the-art MARL algorithms~\citep{de2020independent,yu2022surprising} and recently proposed social learning methods~\citep{ndousse2021emergent}.

%

\textbf{Emergent Collaborative Tool Use.} 
Tool use is a core component of human social learning and our cultural and technological evolution \citep{van2003model,humphrey1976social,boyd2011cultural}. A person creating and leaving a tool in a shared environment facilitates social learning by structuring the environment so others can more easily discover how to use and manufacture tools \citep{henrich2017cultural}.
\methodabbrev{} uniquely presents an opportunity to study how agents share tools in an environment where one agent's progress fundamentally alters the environment to benefit others.

\textbf{Cooperation vs. Competition in Mixed-Motive Settings.} We study how competitive and cooperative pressures within \methodabbrev{} affect social learning, tool sharing, and agent performance. 
The environment has mixed-motive incentives, where agents can benefit both from cooperating (by sharing tools or teaming up against common enemies) and competing (by consuming resources or attacking others). By varying the environment's reward structure, researchers can study how agents balance cooperation and competition and its impact on  social learning  and emergent behaviors.


While we find interesting evidence of shared tool use,
we find little indication that existing methods are effective at using social learning to improve performance on the demanding, long horizon tasks in \methodabbrev{}.
This suggests further research into improving the ability of state-of-the-art MARL algorithms to rapidly acquire skills from others using social learning is needed. \methodname{} provides a new way to study this, and particularly examine how agents benefit from sharing an environment with other intelligent agents who restructure it to be safer and provide shared tools.



\section{Related Works}
\textbf{Multi-Agent Environments.} Many MARL environments (e.g. MADDPG~\citep{lowe2017multi}, Starcraft Multi-Agent Challenge (SMAC) ~\citep{DBLP:journals/corr/abs-1902-04043}, and Melting Pot~\citep{leibo2021meltingpot}) focus on agent cooperation or coordination, but few share our focus on learning from independent agents with long-horizon, hard-exploration tasks.
While Neural MMO is a massively-multiplayer open-ended environment,~\cite{suarez2019neuralmmomassivelymultiagent} mainly investigate how competition drives the emergence of intelligent behavior. We test both competition and cooperation, but primarily focus on studying and improving the ability of agents to learn socially from others and implicitly cooperate.

\textbf{Minecraft Research.}
Many environments based on Minecraft~\citep{minecraft,guss2019minerllargescaledatasetminecraft,mao2021seihaisampleefficienthierarchicalai}, such as Minecraft Building Assistance Game~\citep{laidlaw2024scalably} and VillagerBench~\citep{dong2024villageragentgraphbasedmultiagentframework}, have been created because of its long horizon, sparse rewards, and potential for emergence of complex behavior.
One such case is Crafter~\citep{hafner2021benchmarking}, which 
is designed to be challenging, with achievements representing significant milestones in performance. 
However, because it can only be simulated on a CPU due to being written in pure Python, the benchmark is limited to only 1 million environment interactions.

\textbf{JAX-based environments.}
JAX~\citep{jax2018github} is a Python library that compiles and optimizes array computation and automatic differentiation to run on hardware accelerators such as GPUs and TPUs, enabling speedup of many magnitudes~\citep{rutherford2024jaxmarlmultiagentrlenvironments}.
Several multi-agent environments based on JAX have been proposed recently (e.g. \citet{ruhdorfer2024overcooked,jhainfinitekitchen,hou2024investesg}), but they again focus on cooperation. Of greater interest is Craftax~\citep{craftax2024}, a single-agent JAX implementation of Crafter. 
It offers significant
computational speed improvements, running up to 250 times faster than the Python-native
Crafter, while retaining the core gameplay.
Even in the single-agent setting, current RL methods like global
and episodic exploration struggle to make significant progress on this benchmark, demonstrating its value for driving AI research~\citep{craftax2024}. In this work, we build on Craftax by adapting it for the multi-agent setting. We believe that our \methodabbrev{} environment is the first multi-agent minecraft-like environment entirely written in JAX.

\textbf{Multi-Agent Social Learning.}
It is well-known that acquiring skills by learning from expert agents via imitation learning can massively improve sample complexity vs. random exploration, especially in sparse reward problems (see e.g. \citet{nair2018overcoming}). Recently, there have been efforts to train RL agents to use social learning to \textit{learn to learn} from experts to acquire information that can enable online adaptation to novel environments \citep{ndousse2021emergent,filos2021psiphi}. 
Additionally, \citet{bhoopchand2023learning} proposed a metric to quantify the degree of social learning, which we adapt in this work. A 2010 study by \citet{rendell2010copy} found that copying strategies improved performance in a multi-agent multi-armed bandit tournament. They found that the more agents copied, the better they did in the tournament, with the most successful strategies performing copy on over 90\% of turns \citep{rendell2010copy}. Copying was so effective since each agent was attempting to win and play the most rewarding strategy. Thus, we designed \methodabbrev{} to represent a realistic multi-agent environment in which other self-interested, intelligent agents independently pursue their own goals.


\section{Background}

\textbf{Partially Observable Markov Games.}
The environment is modeled as a partially observable Markov game $M = (S, A, T, R, \Omega, O, \gamma)$ such that $S$, $A$, and $\Omega$ define the environment's joint state, action, and observation space respectively \citep{grupen2022conceptbased}. In each environment interaction, agent $i$ selects action $a_i \in A_i$, yielding the joint action $(a_1, \ldots, a_n)$ for all agents. The environment subsequently transitions to a new state using $T : S \times A \to S$, and produces the corresponding observations consistent with $O : S \times A \to \Omega$. Finally, the reward is calculated according to $R : S \times A \to \mathbb{R}^n$. The goal of each agent's policy $\pi_i : \Omega_i \to A_i$ is to maximize its own expected future discounted reward $\mathbb{E} \left[\sum_{t=0}^\infty \gamma^t r_{ti}\right]$ where $\gamma \in [0,1]$ is the discount factor and $r_{ti}$ is the reward at time $t$ for agent $i$. The set of all agents' policies $\pi = (\pi_1, \ldots, \pi_n)$ is referred to as the joint policy.

\section{Environment Mechanics}
In this work, we extend the \texttt{Craftax-Classic} environment to make it multi-agent. This section describes the game mechanics of the environment.

\begin{figure*}
    \resizebox{\textwidth}{!}{
      \includegraphics{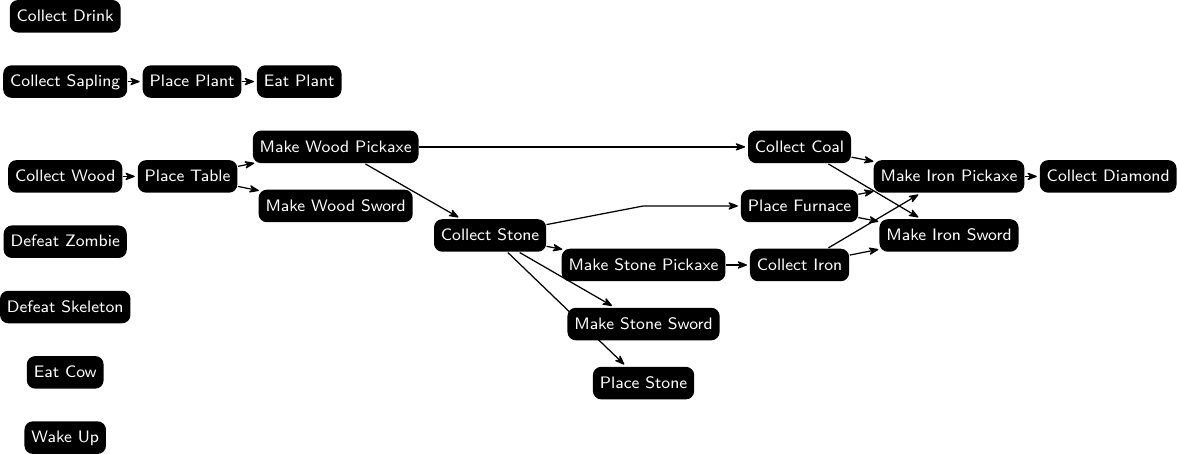}
    }
    \caption{
        Tech Tree of the 22 achievements, 
        with an arrow pointing to a task from each of its prerequisites. Not all prerequisites must be completed if another agent achieves them instead (e.g. an agent can use a table placed by another to make a pickaxe without completing the `Place Table' task). Some prerequisites may also need to be repeated (e.g. Two wood is required to place a table).
    }\label{fig:tree}
\end{figure*}

\textbf{Game Objective.}
The core game is equivalent to \texttt{Craftax-Classic} ~\citep{craftax2024}, a performance-optimized version of Crafter~\citep{hafner2021benchmarking}. Agents complete as many achievements as possible in a
single episode while surviving with positive health by keeping their intrinsics (food, water, energy) positive and avoiding attacks from zombies and skeletons (mobs). Some achievements, such as making a Furnace, require prerequisite tasks such as collecting stones. Other achievements, such as \texttt{EAT\_COW}, can be
completed without prerequisites. Figure~\ref{fig:tree} shows the (technology) tree of the 22 Achievements, structured to balance depth and breadth. Note that achievements are the main comparison method of the performance of different training methodologies.


\textbf{Observations and Actions.}
The obserservation and action space is the same as \texttt{Craftax-Classic}, except at each timestep, the environment provides additional information about other players to each player in the observation space, and operates with one observation and action per player. For more information, see Appendix~\ref{a:obs} and~\ref{a:act}.

\textbf{Rewards.}
Rewards are calculated for each agent individually and returned
as an array with the same length as the number of agents.
Each agent would get a reward of +1 if an achievement was achieved during an
environment step. Additionally, to encourage agents to preserve their health,
a reward of 0.1 times the change in the player's health is added. 


\textbf{Player Constraints.}
In \methodabbrev{}, no two agents can place a block or perform a \texttt{DO}
on the same space. If two or more players attempt to do so during same time step, one random player will complete the action, while the other players' actions are ignored. This prevents players from exploiting the game to acquire duplicate resources from the same block, and more closely models real-world dynamics.


\textbf{Mob Logic.}
We update the mob logic to only consider the nearest player when determining the mob's next move, using distance from this player to determine whether to spawn or despawn.


\textbf{Measuring Tool Use.}
We modified the environment to track which agent placed a block, and then tracked how often each agent used its own crafting table versus one placed by another agent.

\section{Quantifying Social Capabilities}
\label{sec:capabilities}

In our experimental setup, we investigate three facets of social intelligence: 1) social learning, 2) the emergence of collaborative tool use, and 3) cooperation and competition. We designed experiments to explore each of the following facets. 

\textbf{Social Learning.}
Agents which are effective social learners should be able to learn more effectively when there is an expert present in their environment demonstrating highly rewarding behavior. Therefore, we pre-train an expert policy in the environment, and 
investigate a setting where we have an agent training with other pre-trained experts, updating only the agent every rollout.
Since we want to measure social learning, all agents should benefit equally from tools placed by the expert. The experts are placed normally in the environment for all instances, able to perform the same actions as the player, but are made invisible to the player's observations when they are ``removed''. In order to quantify whether agents are able to acquire knowledge from experts via social learning, and maintain that improvement once the demonstrator has departed, we measure the cultural transmission score first introduced in~\cite{bhoopchand2023learning} in order to compute how much having an expert present improves performance. Cultural transmission is defined as
\[
CT := \frac{1}{2} \frac{A_\text{full} - A_\text{solo}}{E} + \frac{1}{2} \frac{A_\text{half} - A_\text{solo}}{E},
\]
where $A_\text{full}$ is the agent's score trained with the expert, $A_\text{solo}$ is the agent's score trained without observing the expert, and $A_\text{half}$ is the agent trained initially with the expert, but with the expert removed from its observations after 50 time steps in every episode. $E$ is the performance of the expert.  We use this measure to quantify the degree of social learning of different methods in \methodabbrev{}.

\textbf{Emergent Collaborative Tool Use.}
We begin by comparing the average performance of agents in the multi-agent environment with the performance of their single agent counterpart. Unlike in the previous setting, all agents are learning from scratch and there are no experts. Thus, improvements in 
performance due to having multiple agents vs. a single-agent could be due to the emergence of cooperation, social learning, or potentially through shared use of tools.
To test this, we modified the environment to track which player placed down each crafting table and furnace and compared how often an agent used its own tools versus another agent's. Should agents that use others' resources perform better, this could present collaborative behavior as an area of further research.
    

\textbf{Cooperation and Competition.}
We investigate the behavior of agents in the open-world environment to observe if they exhibit cooperative behavior, or if they deem it better to perform actions independently. To incentivize cooperation, we train agents on a shared collective reward. To incentivize competition, we introduce a $+1$ reward for attacking other agents, and a $-0.5$ reward for being attacked. 
We also track the agents' proximity by measuring the fraction of time players are within view of each other. We experiment with rewarding agents for maintaining higher proximity to assess impact on performance. A correlation between higher proximity and higher scores would imply the environment implicitly incentivizes cooperation, perhaps through sharing tools or because the group offers protection from mobs. Alternatively, a correlation between greater distance from other agents and higher scores could imply that competing for resources poses a significant challenge. 


\textbf{Implementation of Baseline Methods.}
We used Independent PPO \citep{de2020independent,yu2022surprising}, a state-of-the-art MARL algorithm 
which trains all agents independently with PPO with no additional information sharing mechanisms. We evaluated all of the agents simultaneously each step, and independently
applied PPO updates after each environment batch.
The PPO algorithm is based on CleanRL~\cite{huang2022cleanrl},
\textit{Recurrent PPO in JAX}~\citep{pramanik2022}, and
PureJaxRL~\cite{lu2022discovered}. For the social learning algorithm, we chose to use a version where an auxiliary loss is added to the loss for each agent's loss~\citep{ndousse2021emergent}. All experiments were run with 4 agents with 100 million steps on an Nvidia L40 GPU, which allowed hundreds of environments to be run in parallel.

\begin{wraptable}[17]{R}{0.5\textwidth}
    \vspace{-0.5cm}
    \centering
    \caption{Performance of Agents under different scenarios. In solo, the player cannot see the expert. For half, agents can see the expert for 50 timesteps each episode\@. (aux) agents are trained with a social auxiliary loss, following \citet{ndousse2021emergent}. The cultural transmission score was $-0.056 \pm 0.083$ and $-0.010 \pm 0.080$ with and without auxiliary loss respectively, implying that agents do not benefit from social learning in MAC.}
    \begin{tabular}{l l}
         \toprule
         Scenario & Mean Achievements \\
         \midrule
         Solo & $15.44 \pm 1.28$ \\
         w/ Expert & $14.44 \pm 1.37$ \\
         w/ Expert (aux) & $15.44 \pm 1.35$ \\
         w/ Expert half & $14.68 \pm 1.31$ \\
         w/ Expert half (aux) & $15.12 \pm 1.29$ \\
         Expert Score & $15.84 \pm 1.78$ \\
         \bottomrule
    \end{tabular}
    \label{tab:expert-performance}
\end{wraptable}
\section{Experimental Results}

\subsection{Social Learning}

\begin{figure}
    \includegraphics[width=\linewidth,trim={0 0 0 5cm},clip]{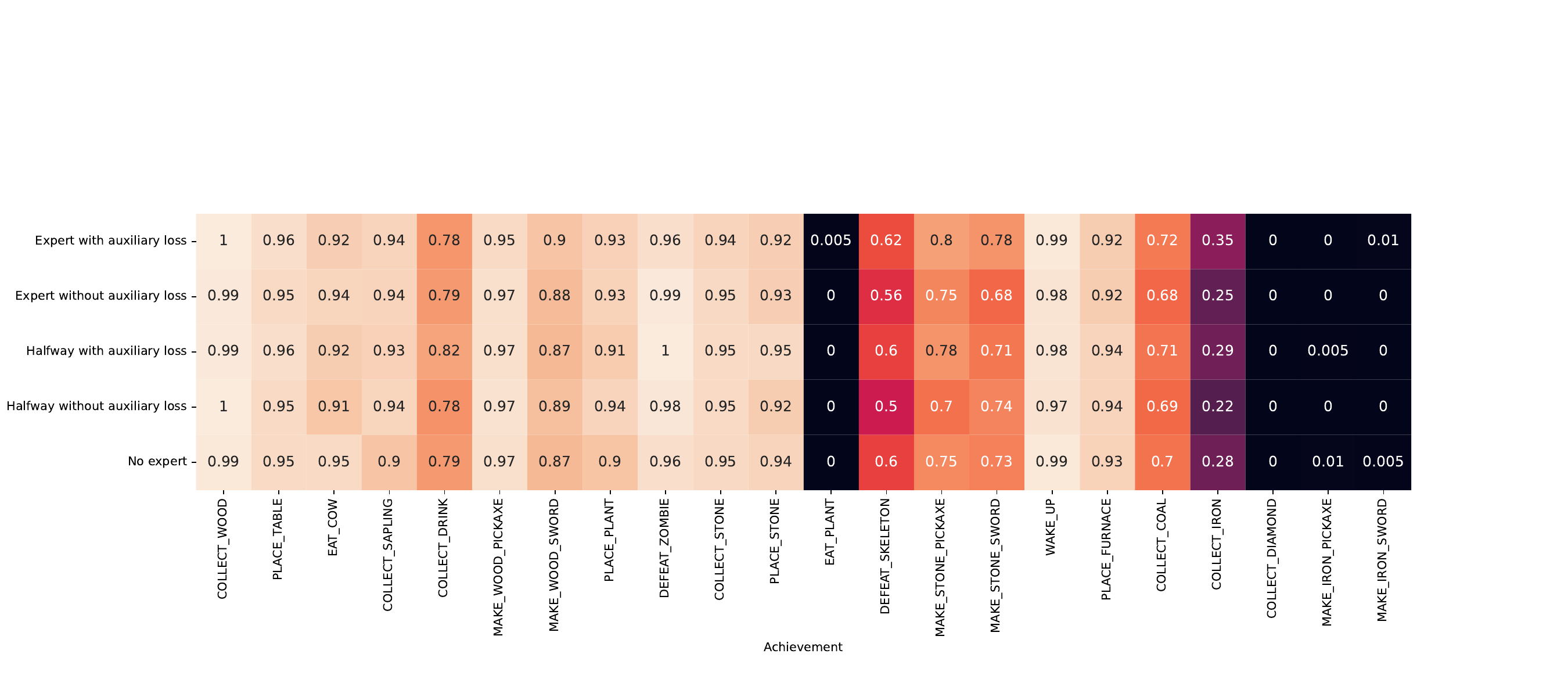}
    \caption{Achievement probabilities after training with the expert during the entire episode and halfway in the same episode, with and without social auxiliary loss.}\label{fig:aux_loss_heatmap}
    \vspace{-0.2cm}
\end{figure}

We initially trained experts using IPPO. Subsequently, we placed those experts in the same environment as a new agent, and at each step, we only updated the new agent. We repeated this experiment but removed the observations of the experts halfway after 50 steps into the environment. Furthermore, we tested a scenario where the agent could not observe the expert at all as a control. As per Table~\ref{tab:expert-performance}, there was no discernible improvement in observing the expert, and we achieved almost identical average achievements per episode. For a more specific achievement breakdown, Figure~\ref{fig:aux_loss_heatmap} illustrates the probability of attaining each individual achievement per episode.

Given that the expert used to train the agents achieved $15.84\pm1.78$ per episode, this yields a cultural transmission of about $-0.056 \pm 0.083$ without auxiliary loss and about $-0.010 \pm 0.080$ with auxiliary loss. Since factoring in error the cultural transmission hovers around 0, this suggests that the presence of experts in our current setup does not help agents learn faster. We investigate this in later sections by verifying whether agents are within view to each other, which is vital for copying.

\begin{wrapfigure}[18]{R}{0.5\textwidth}
\centering
\vspace{-1em}
\includegraphics[width=\linewidth]{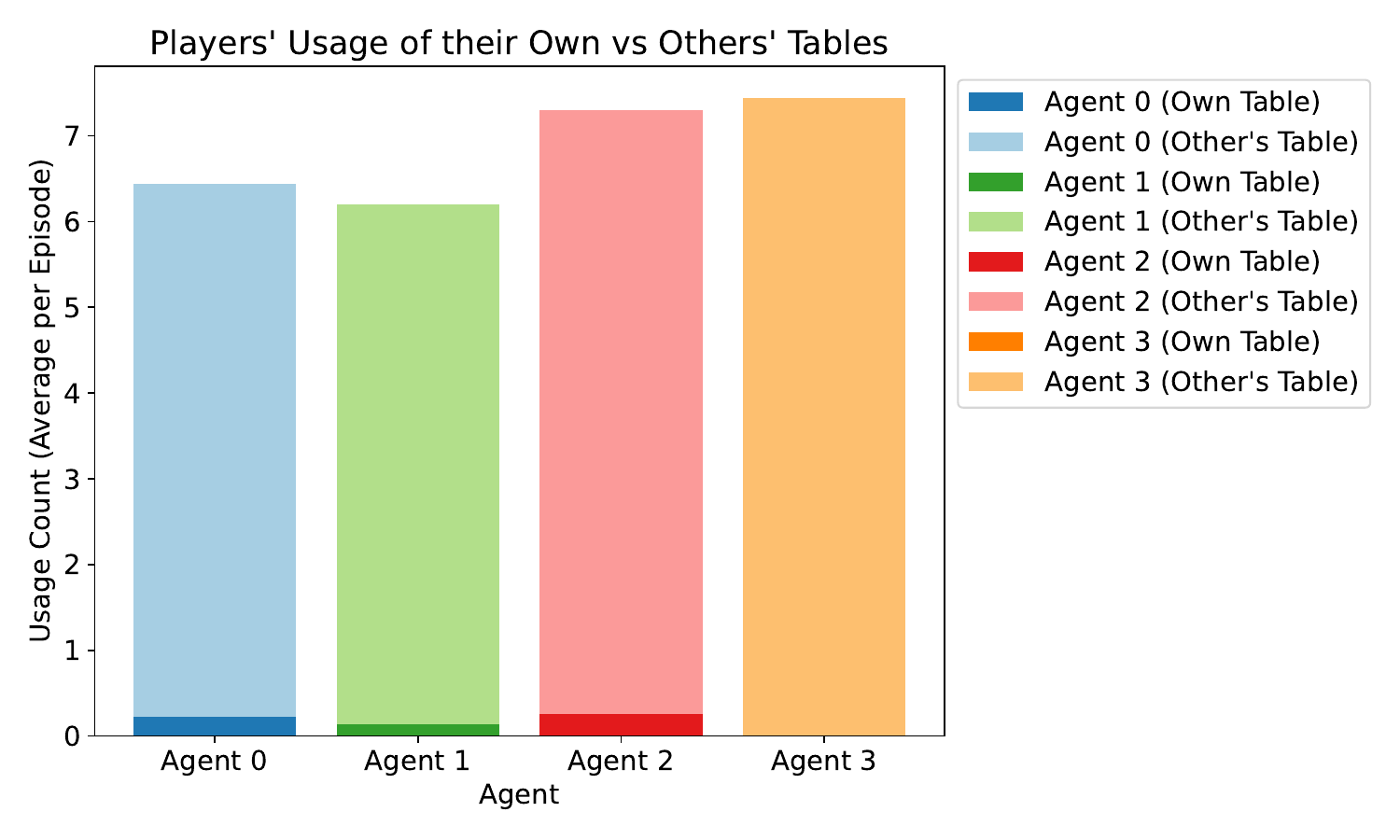}
\caption{Count of occurrences where players use their own table (bottom bar) versus when players use another players' table (upper bar). The probability that a player uses their own table is less than random chance.}
\label{fig:graph4}
\end{wrapfigure}




\subsection{Emergent Collaborative Tool Sharing}

\begin{figure}
    \centering
    \includegraphics[width=0.8\linewidth,trim={0 0 0 0.4cm}]{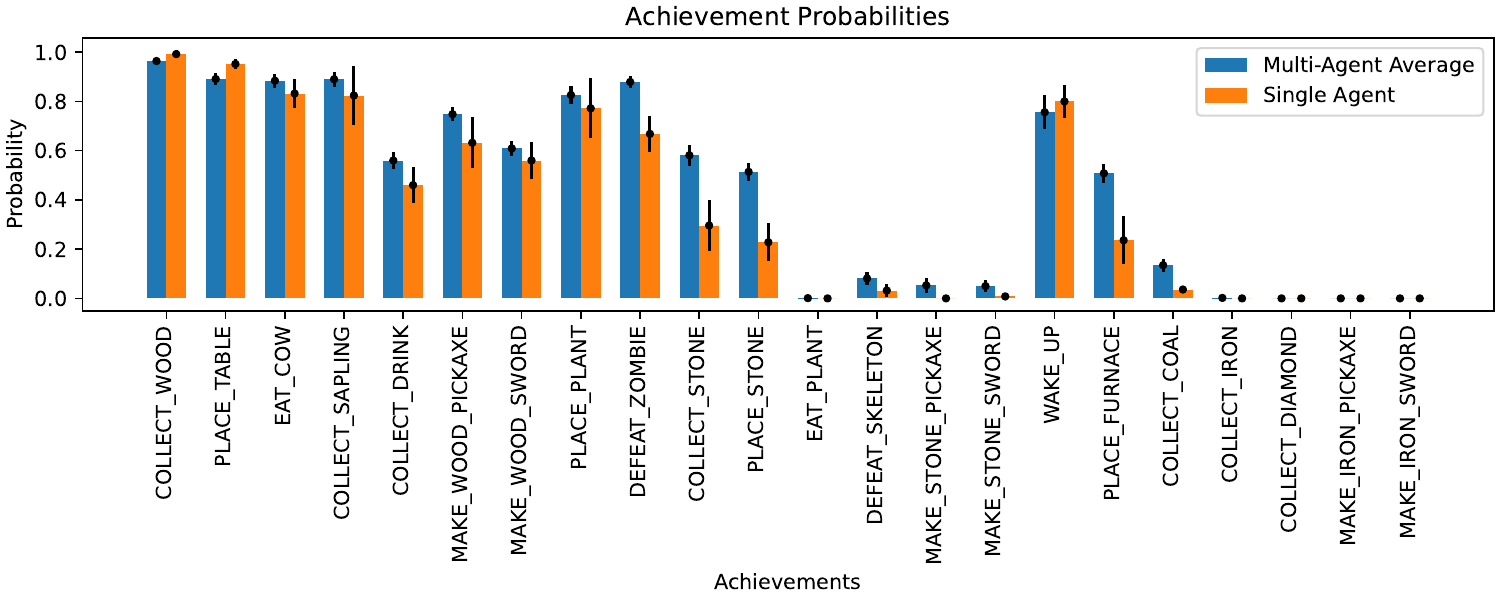}
    \caption{Achievement Graph of Single Agent and Multi Agent Average with fixed timesteps. Achievements are averaged over 5 training runs, sampled over 50 episodes each, and in the multi-agent case, it is averaged over all agents. Error bars are standard deviation.}\label{fig:single_vs_multi}
    \vspace{-0.2cm}
\end{figure}

The first part of this investigation consists of comparing single and multi-agent performance on the achievements in the game. As seen in Figure~\ref{fig:single_vs_multi}, we notice
that the single agent version and the multi-agent version have similar scores, but in the multi-agent environment, agents had higher average probability of reaching the more difficult achievements, such as placing furnaces and collecting coal.
On the other hand, training with an expert that can be observed or training from scratch (see appendix ~\ref{a:aux_performance}) did not offer any improvements compared to training multiple independent agents.

Thus, a better multi-agent performance does not necessarily imply that agents benefit from social learning, even if other agents develop skills that could be useful to learn from. Instead, higher multi-agent performance can be attributed to agents sharing tools. Since agents can use other agents' tools, they may be able to skip unnecessary prerequisites specified in Figure~\ref{fig:tree}. For example, if an agent lacking the materials for a crafting table finds one placed by another agent, it can more quickly reach other achievements. Additionally, repeating an achievement does not provide further reward, so it can be more beneficial to find an existing crafting table rather than crafting another.

In Figure~\ref{fig:graph4}, we show the number of times agents used a crafting table that they themselves built vs. the number of times they used a table built by another agents. According to the figure, when using a table, agents had a less than 10\% chance of using their own table, which is less than the probability of randomly picking an agent's table with equal chance. This possibly suggests that agents tend to (perhaps unintentionally) favor other agents' tools. Thus, even if the agents are self-interested, they unknowingly assist each other simply by modifying the shared environment. 

\subsection{Cooperation versus Competition}
\begin{figure}
        \centering
        \includegraphics[width=0.8\linewidth,trim={0 0 0 0.7cm}]{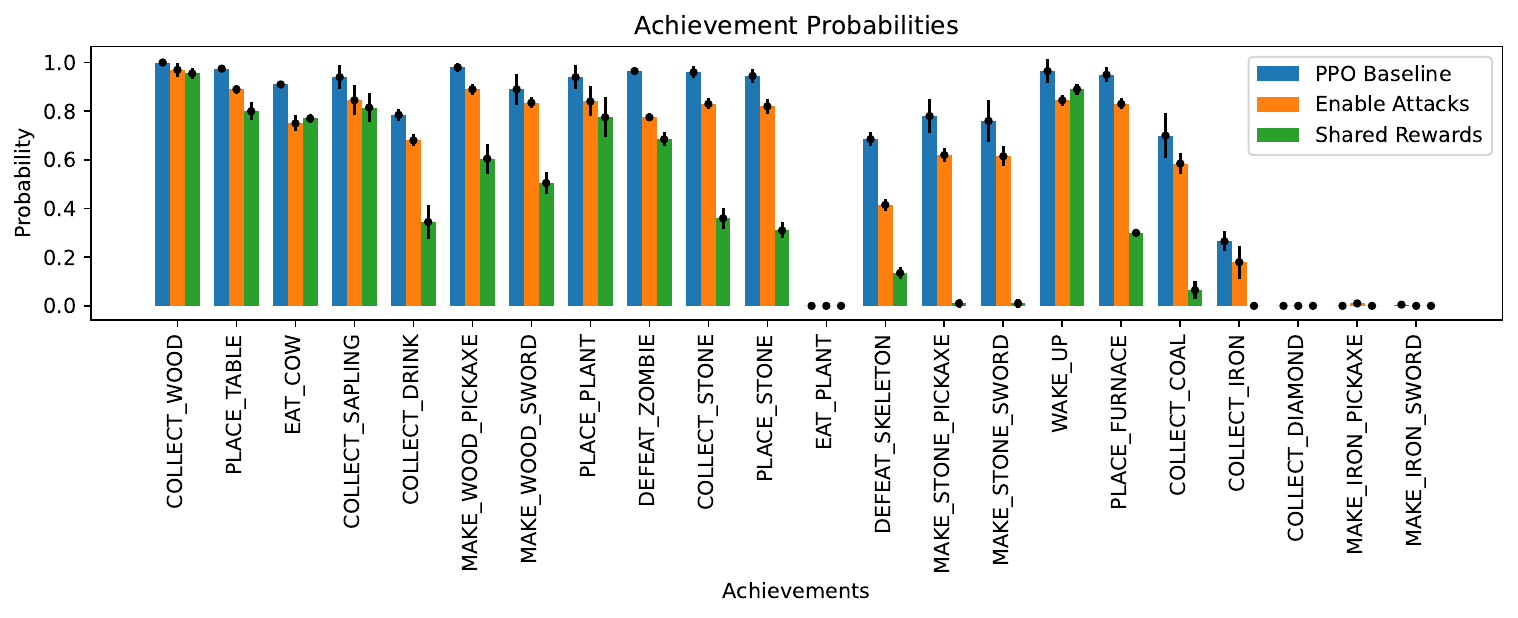}
        \caption{Achievement probabilities of baseline, enabling attacks, and sharing the total rewards. Timesteps are not fixed.}\label{fig:reward_cooperation_competition}
\end{figure}

As described in Section \ref{sec:capabilities}, we experiment with increasing both competitive pressures (enabling agents to attack each other), and cooperative incentives (training on a shared reward). 
As shown in Figure~\ref{fig:reward_cooperation_competition}, the total achievements attained in both settings decreased from the baseline IPPO with independent rewards scenario. It is likely that the shared cooperative reward was harder to optimize because it provided a noisier signal about which of each agent's actions led to higher rewards \citep{foerster2018counterfactual}. However, agents could complete basic tasks. In the competitive setting, achievements also decreased, suggesting that in this environment, increased competition led to lower performance, rather than driving the emergence of more complex behavior, as suggested by \citet{suarez2019neuralmmomassivelymultiagent}.

\begin{figure}
    \centering
    \begin{subfigure}[t]{0.32\linewidth}
        \centering
        \includegraphics[width=\linewidth]{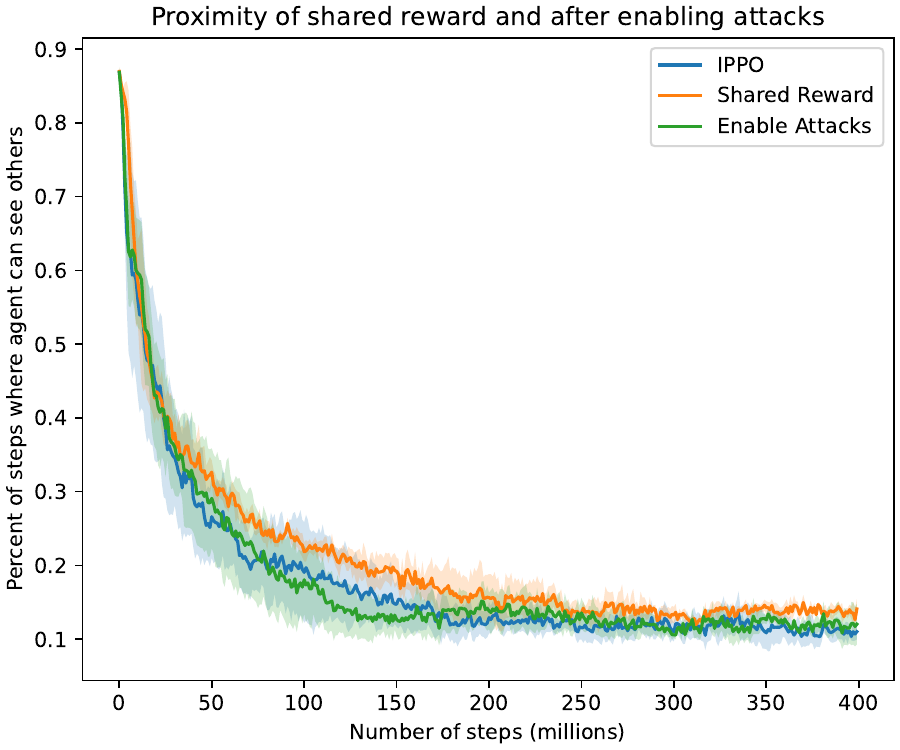}
        \caption{Proximity in competitive and cooperative settings.}
        \label{fig:proximity_shared_attack}
    \end{subfigure}
    \hfill
    \begin{subfigure}[t]{0.65\linewidth}
        \centering
        \includegraphics[width=\linewidth]{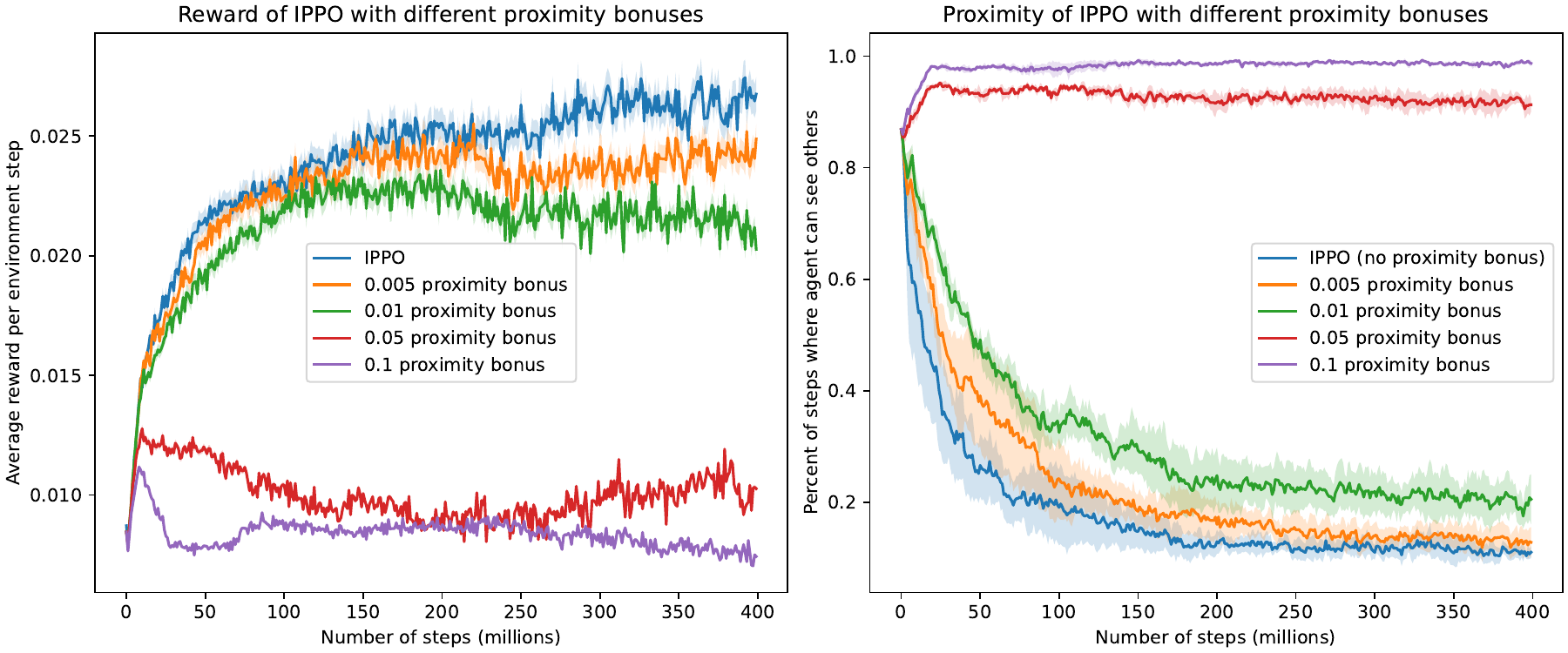}
        \caption{Average reward and proximity of players with different levels of proximity bonus.}
        \label{fig:proximity}
    \end{subfigure}
    
    \caption{Training curves showing both reward and proximity (the fraction of environment steps where agents can see each other). We investigate how altering competitive vs. cooperative incentives result in changes to proximity and performance. }
\end{figure}

Figure~\ref{fig:proximity_shared_attack} shows one possible explanation, plotting the proximity agents maintain from each other in the default, competitive, and cooperative setting. Predictably, agents maintain greater proximity in the cooperative setting, and less in the competitive setting. The fact that greater competition drove agents further apart, and also resulted in lower achievements, suggests an interesting interpretation.


We know from studies of human social learning (e.g. \citep{henrich2015big}) that for people to observe and learn from one another, they must maintain close proximity. Therefore, we hypothesized that a lack of proximity between the agents in \methodabbrev{} could explain why techniques that have previously proven useful for enabling social learning in other environments (such as the social auxiliary loss proposed by \citet{ndousse2021emergent}) did not lead to higher performance, even in the presence of experts. As shown in Figure~\ref{fig:proximity}, without any explicit reward, agents do not stay close to each other in \methodabbrev{}, which provides support for this hypothesis. 

To counteract this, we tried adding a proximity bonus to the reward in order to incentivize players to stay close to each other. While increasing the proximity bonus increased the proximity (as expected) it decreased the rewards from the environment. In Figure~\ref{fig:proximity}, the additional proximity bonus allowed players to learn to follow each other, but this resulted in lower total rewards and achievements. The lack of proximity as the training went on can explain why adding a social auxiliary loss did not improve compared to IPPO, since it is based on predicting information about agents within each agent's field of view. We hypothesize that because current social learning techniques do not yet enable rapidly acquiring skills from other observable agents, staying near other agents does not pay off in terms of skill discovery, but proves costly because it reduces exploration and resource collection. 

\section{Discussion and Conclusion}
We propose the first multi-agent adaptation of Craftax, presenting significant exploration, long-term planning, and reasoning challenges. 
\methodabbrev{} enables rapid iteration and improvement of the social intelligence of AI agents. 
We conduct initial tests of state-of-the-art MARL algorithms with respect to three capabilities: 1) social learning, 2) tool sharing, and 3) cooperation and competition. We find that agents learning in a multi-agent world in the presence of a pre-trained expert do not perform significantly better than agents training  alone, indicating they have little ability to acquire skills from other intelligent agents via social learning. However, the multi-agent environment does provide some boost to acquiring more complex skills because other agents modify the environment, enabling agents to share crafted tools. We test the effects of both cooperation and competition in the environment, finding that these incentives can shape the proximity agents maintain to each other and in turn their performance. 
Our results suggest that significant research in developing better social learning algorithms is needed.
We hope that this benchmark will accelerate this research and bring us a step closer toward socially intelligent agents that can rapidly acquire skills and generalize to new environments by learning from others. 

\textbf{Future Work.}
While it is challenging for agents to perform well in this environment in a small number of interactions, \citet{craftax2024} produced a model that consistently completed around 90\% of achievements after training for a billion interactions in the \texttt{Craftax-Classic} environment, but lacked good sample efficiency.
However, we believe that if agents could build on skills acquired from other agents via social learning, sample complexity could be significantly reduced.

\medskip
\bibliography{references}
\appendix

\section{Agent Observations}\label{a:obs}
The environment offers both Symbolic and Pixel environments. The more human-readable Pixel
environment directly shows the game in terms of RGB pixels (see Figure~\ref{fig:game}), while the Symbolic environment expresses the game state in a more
compact way that is easier to process for machines. Most experiments used the Symbolic environment as it requires a smaller neural network, less training, and faster iteration.
If needed, the observations can be modified relatively easily by modifying the rendering function.
Returning an array of observations also makes it possible to implement other multi-agent algorithms which allow agents to share observations as a way of enabling social learning through post-processing of the observations.

\subsection{Symbolic Environment Observations}
The following are the observations in the Symbolic Environment.
\begin{itemize}
    \item 21 $7 \times 9$ grids, one-hot encoded, which show surrounding objects or mobs around the player, who is
          always in the center of the grid. Each grid indicates different objects or mobs.
          A $1$ in a position means that there is a specific mob or object, and $0$ means there is no
          mob or object of a specific type at that location. Observed mobs include zombies, skeletons, cows, arrows, and other players.
    \item An entry for each inventory of the player, 0.0 meaning that the
          player does not have a specific item in the inventory, and 0.9
          meaning that the player has 9 items of that type in their inventory
    \item An entry for each of the player intrinsic (health, food, drink,
          energy). 0.0 means that the player does not have any of that
          intrinsic, and 0.9 means that the player has the max of that
          intrinsic.
    \item One hot encoding for each of the 4 possible player directions
    \item Light level of the environment
    \item Whether the player is sleeping
    \item Whether the player is alive
\end{itemize}
We also offer a setting for players to observe each other, in which case the following items are also visible for each player if the player is within view:
\begin{itemize}
    \item One-hot $7\times 9$ map, with a 1 in the position of the player
    \item Inventories of the player
    \item Intrinsics of the player (health, food, drink, and energy)
    \item One-hot encoding of the player's direction
\end{itemize}

\section{Agent Actions}\label{a:act}
This environment implements a discrete action space, which can be used with many modern deep RL algorithms.
At each time step, each player picks one of 17 actions, which includes movement actions,
the \texttt{DO} action for interacting with objects, crafting actions, or \texttt{NOOP}, which performs no action.
\begin{itemize}
    \item \texttt{NOOP}: Don't perform any action
    \item \texttt{LEFT}, \texttt{RIGHT}, \texttt{UP}, \texttt{DOWN}: Move actions. This is equivalent to a \texttt{NOOP}
          if there is a mob, tree, stone, plant, or water blocking the way.
    \item \texttt{DO}: The \texttt{DO} action performs a different action depending on the
          block that the player is facing.
          \begin{itemize}
              \item Grass: The player will try to collect saplings.
              \item Stone, coal, iron, or diamond block: the player will try to
                    mine that block.
              \item Tree: The player will mine for wood.
              \item Water: The player will drink and replenish the drink
                    intrinsic.
              \item Cow: The player will eat the cow and replenish the hunger
                    intrinsic.
              \item Plant: If ripe, the player will eat fruits on the plant and
                    replenish the hunger intrinsic.
              \item Zombie or Skeleton: The player will try to attack.
              \item Player: Attack player, gaining 0.5 health, and removing 1 health from the victim. This is only enabled in the setting where we explore cooperation and competition.
              \item Other blocks will be equivalent to a \texttt{NOOP}.
          \end{itemize}
    \item \texttt{SLEEP}: If the player's energy level is less than 9, begin sleeping.
    \item \texttt{PLACE STONE}, \texttt{PLACE TABLE}, \texttt{PLACE FURNACE}, \texttt{PLACE PLANT}: Placement actions
          that will place the specified item if the player has the required inventory.
    \item \texttt{MAKE WOOD PICKAXE}, \texttt{MAKE STONE PICKAXE}, \texttt{MAKE IRON PICKAXE},
          \texttt{MAKE WOOD SWORD}, \texttt{MAKE STONE SWORD}, \texttt{MAKE IRON SWORD}:
          Player will create the specified items given that the conditions are met.
\end{itemize}
\section{Handling of Dead Players.}
During the course of a single episode, some agents will inevitably die before other agents.
In this implementation, the environment will continue to run until all agents have died, either through depletion
of player health or running into lava. During the course of each episode, all agents
but each player will be aware of whether they are dead as a part of their observations. The environment
changes the actions to NOOP, and dead players
do not gain or lose health, and are no longer visible to other players. Therefore, models should be able to learn that no further achievements are possible once dead.

One potential issue
with this current implementation is that agents that perform better will likely last longer,
causing it to have more opportunities to train compared to agents that die earlier. This could allow those agents to perform even better relative to other agents, causing the training curve to diverge. However, this behavior was not observed when training for only 100 million environment interactions.

In order to keep the comparisons fair, when we ran experiments comparing a single agent compared to multiple agents, we fixed the timesteps for each episode, increasing the episode length as training progressed. However, other experiments comparing different instances of multiple agents did not use fixed timesteps.
\section{Training Parameters}
The MLP policy had two hidden layers with size 512 for both the actor and critic networks.
The LSTM policy had a shared network with hidden layers of size 512 and 256, connected
to an LSTM block with 256 features. This is connected to separate actor and critic networks each with a hidden layer of size 256.

For both instances, we used the same parameters as the CleanRL~\cite{huang2022cleanrl} defaults, but with 400 parallel environments, 8 minibatches, 500 steps per batch,
and no learning rate annealing.

\section{Comparing LSTM and Non-LSTM performance}
\begin{figure*}
    \centering
    \begin{minipage}{.5\textwidth}
        \centering
        \includegraphics[width=1\textwidth]{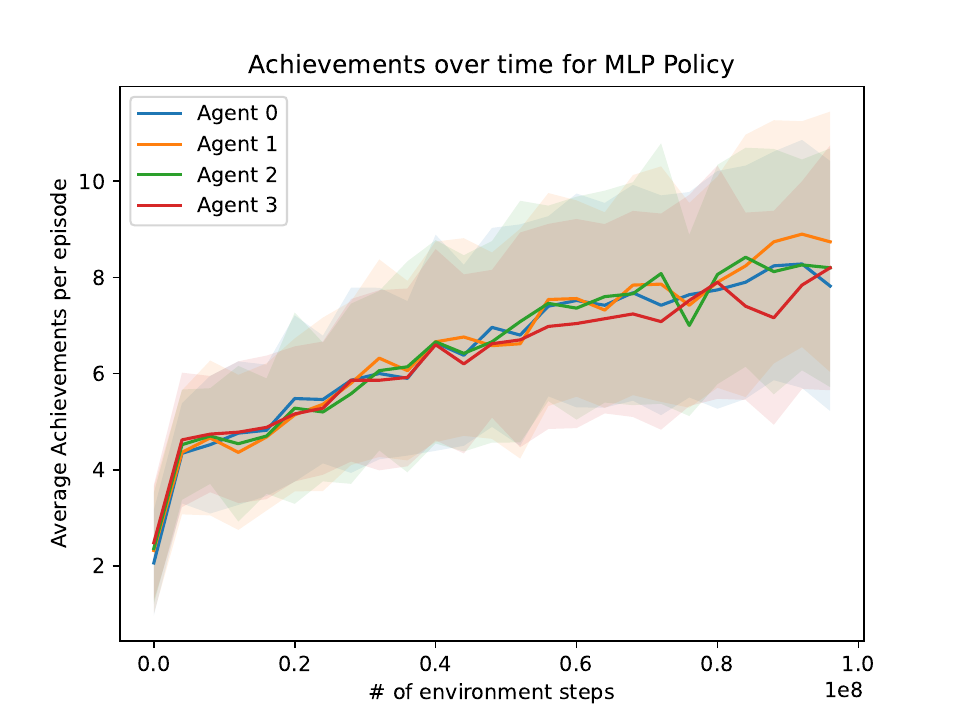}
        \captionsetup{width=0.9\textwidth}
        \caption{
            PPO Achievements during the course of training for 100 million environment interactions with MLP policy.
            The error area represents standard deviation.
        }\label{fig:graph1}
    \end{minipage}%
    \begin{minipage}{.5\textwidth}
        \centering
        \includegraphics[width=1\textwidth]{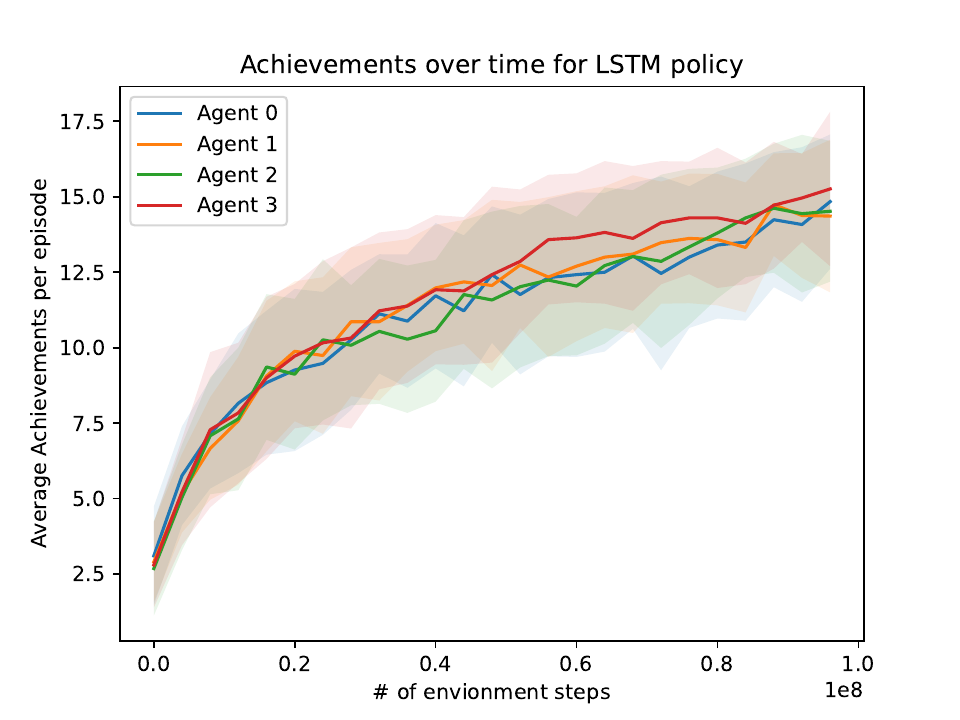}
        \captionsetup{width=0.9\textwidth}
        \caption{
            PPO Achievements during the course of training for 100 million environment interactions with LSTM policy.
            The error area represents standard deviation.
        }\label{fig:graph2}
    \end{minipage}
\end{figure*}

Figures~\ref{fig:graph1} and ~\ref{fig:graph2} were generated by using the mean and standard deviation of the achievements in one episode over 50 random starting environments as all agents had a similar number of achievements per episode throughout training.

Using snapshots of the models during training, we ran each snapshot over a random distribution of environments and averaged the achievements. Comparing Figures~\ref{fig:graph1} and~\ref{fig:graph2}, the LSTM performed better than the model with only linear and activation layers due to the environment being only partially observed. While replaying the episodes, we noticed that the non-LSTM version would occasionally get
stuck if the agent's position was mostly surrounded by objects, while the LSTM version would try a variety of unique
techniques in order to get unstuck.

\section{Performance of Social Auxiliary Loss}\label{a:aux_performance}
\begin{figure}
    \centering
    \includegraphics[width=0.8\linewidth]{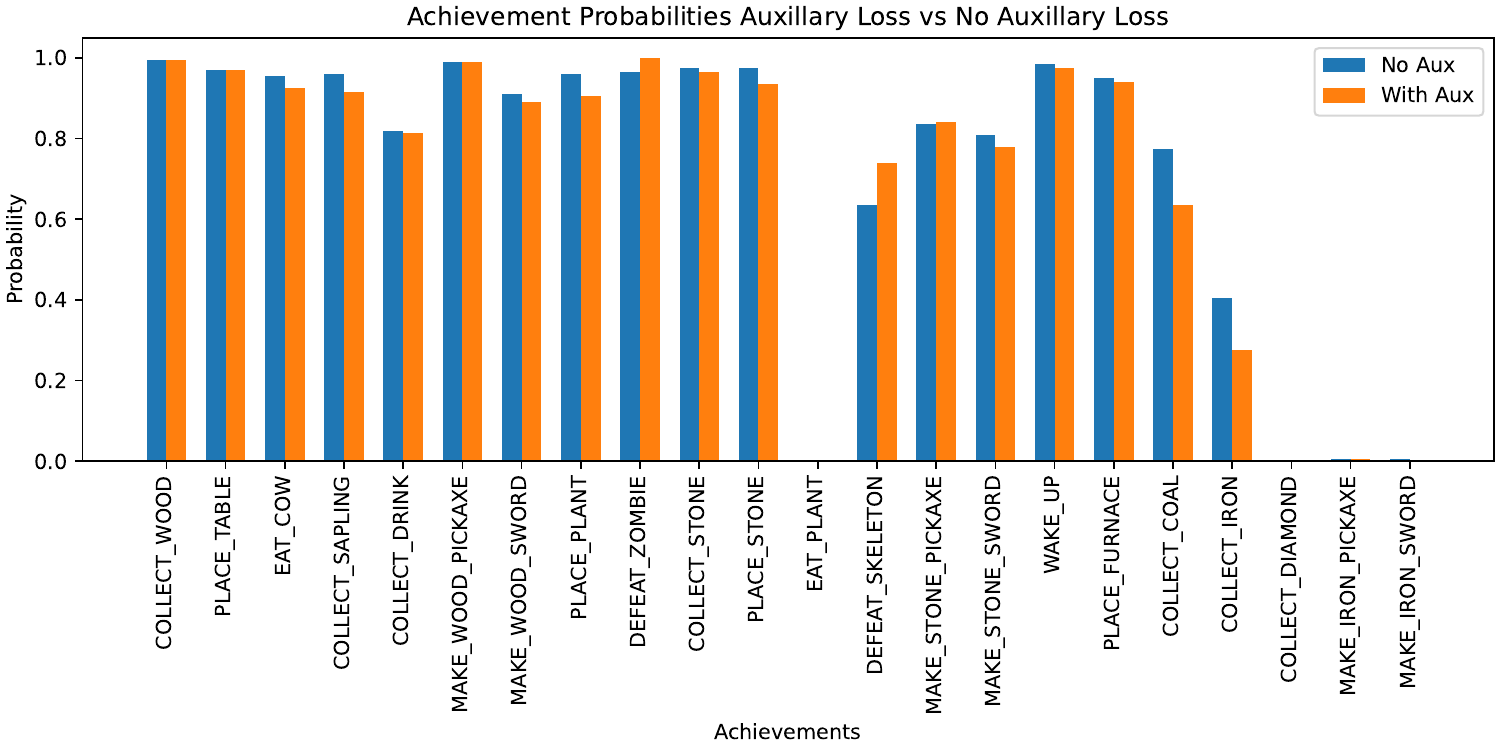}
    \caption{Achievement probabilities after training with four agents, comparing training with and without social auxiliary loss, sampled over 50 episodes each.}\label{fig:graph3}
\end{figure}

We compared agents trained with IPPO with agents trained with an additional auxiliary loss, both of which were trained from scratch and none were expert agents with extra experience. 

As seen in Figure~\ref{fig:graph3}, having the auxiliary loss did not help improve the agents' learning abilities, and in almost all of the cases, adding the auxiliary loss made the agents do worse. 

\begin{figure}
    \centering
    \includegraphics[width=0.8\linewidth]{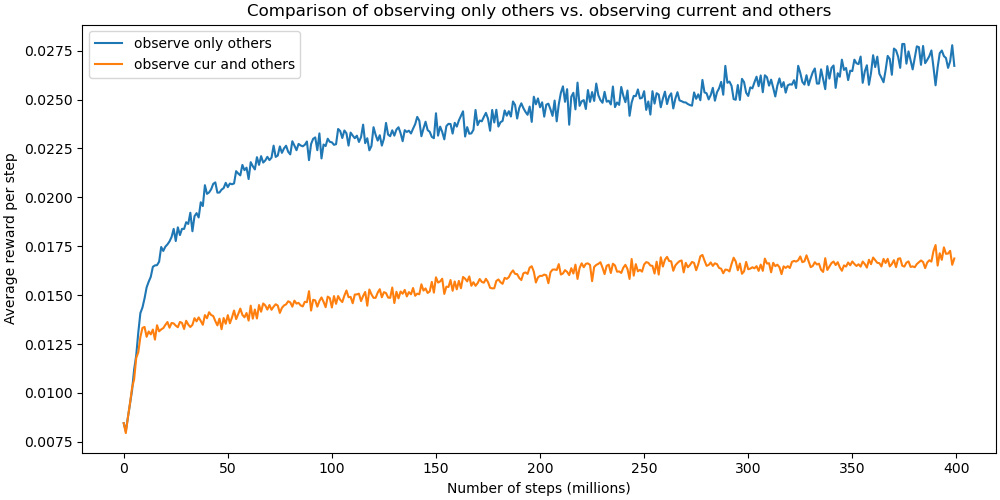}
    \caption{Performance of agents trained with social auxiliary loss predicting only other players versus predicting current player's observations (include the map) and other players.}\label{fig:observe_comparison}
\end{figure}

We tested two versions of Social Auxiliary loss, one where the player predicted both its own map and other player statistics, and one where the player only tried to predict data corresponding to other players.
We found that the performance of the agents is significantly impacted if the agent tries to predict its own map in addition to the positions and inventories of other players as seen in~\ref{fig:observe_comparison}, so in our experiments with the expert, we did not add an auxiliary loss for predicting a player's own features.

\section{Visualization of Social Auxiliary Loss}

\begin{figure}
    \includegraphics[width=\linewidth]{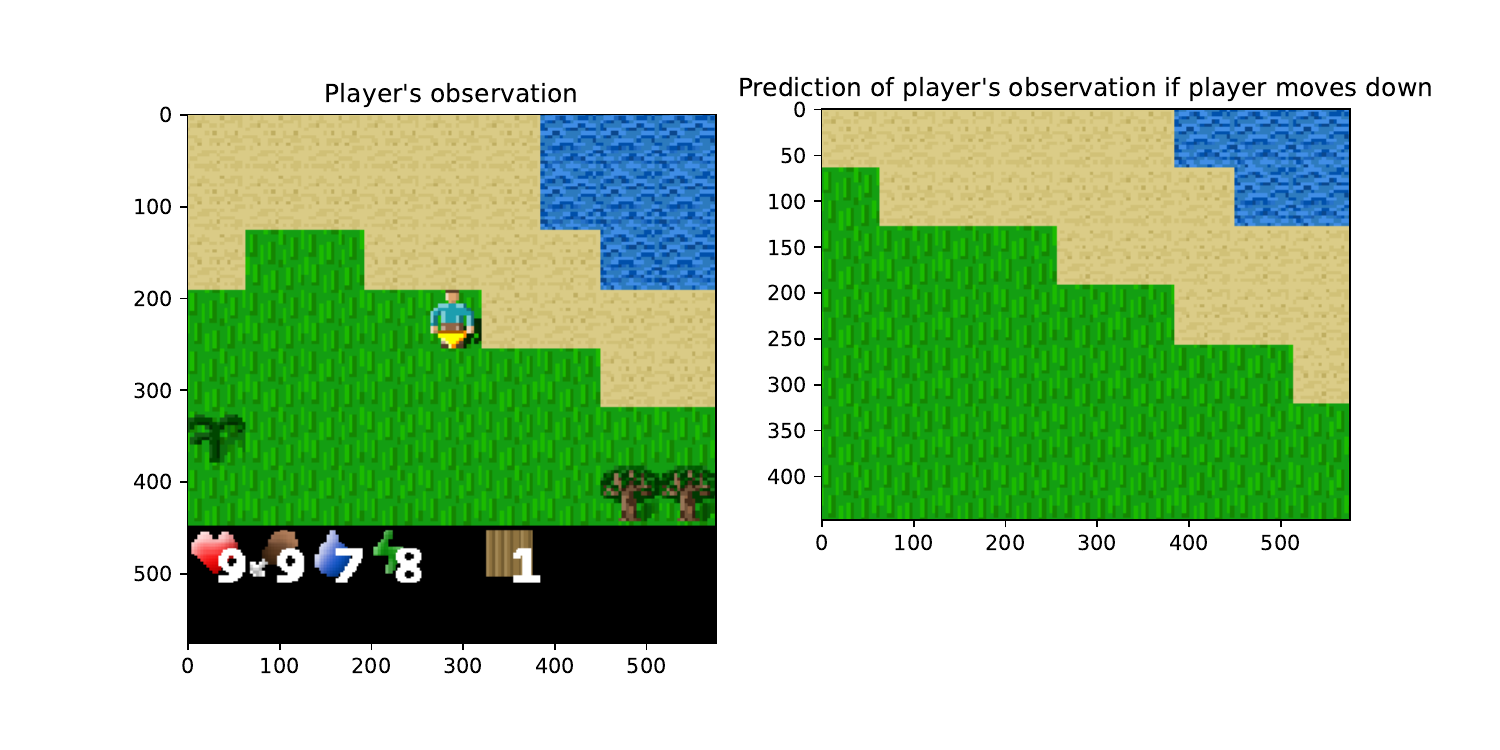}
    \caption{Visualization of auxiliary net's prediction of their surrounding blocks if the player moves down.}\label{fig:pred_obs}
\end{figure}

\begin{figure}
    \includegraphics[width=\linewidth]{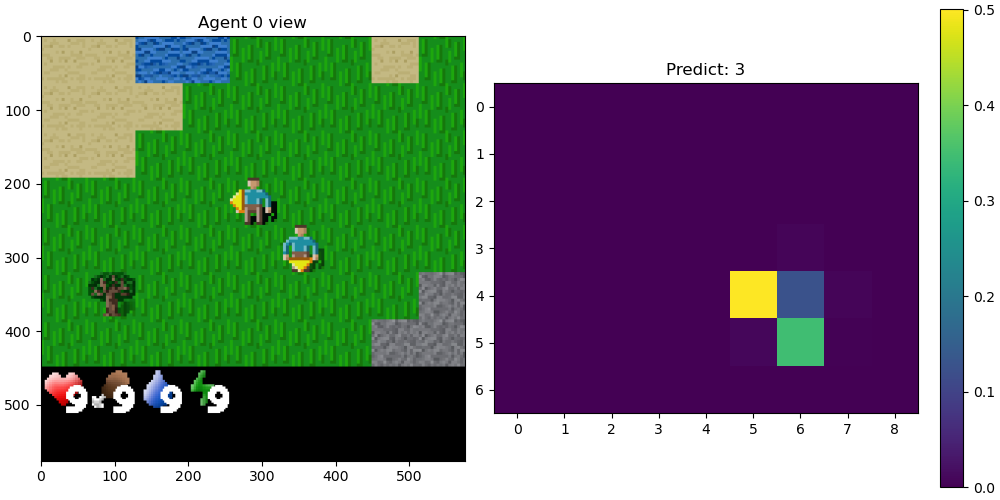}
    \caption{Visualization of Agent 0 and Agent 0's prediction of the future position of Agent 3, given that Agent 0 will move left.}\label{fig:aux_predict}
\end{figure}

In order to verify whether the auxiliary loss first introduced in ~\citep{ndousse2021emergent} is working correctly, we visualized both the agent's prediction of the future state of their own map, as well as positions of other players. In Figure~\ref{fig:pred_obs}, we observe that the agent is able to roughly predict that if the agent moves down, the sand will shift upwards. In Figure~\ref{fig:aux_predict}, we observe that the agent understands that if it moves left, the other agent will be shifted to the right, and it also predicts that the other agent is likely to move left or down.

\end{document}